\documentclass{article}
\usepackage{ijcai24}

\usepackage{times}
\usepackage{soul}
\usepackage{url}
\usepackage[hidelinks]{hyperref}
\usepackage[utf8]{inputenc}
\usepackage[small]{caption}
\usepackage{graphicx}
\usepackage{amsmath}
\usepackage{amsthm}
\usepackage{booktabs}
\usepackage{algorithm}
\usepackage{algorithmic}
\usepackage[switch]{lineno}

\title{BADGE: BADminton report Generation and Evaluation with LLM}

\author{
Shang-Hsuan Chiang
\and
Lin-Wei Chao
\and
Kuang-Da Wang
\and
Chih-Chuan Wang
\and
Wen-Chih Peng
\affiliations
Department of Computer Science, National Yang Ming Chiao Tung University, Hsinchu, Taiwan
\emails
andy10801@gmail.com,
william09172000@gmail.com,
gdwang.cs10@nycu.edu.tw,
wangcc@nycu.edu.tw,
wcpeng@cs.nycu.edu.tw
}

\begin{document}

\maketitle

\begin{abstract}

Badminton enjoys widespread popularity, and reports on matches generally include details such as player names, game scores, and ball types, providing audiences with a comprehensive view of the games. However, writing these reports can be a time-consuming task. This challenge led us to explore whether a Large Language Model (LLM) could automate the generation and evaluation of badminton reports. We introduce a novel framework named \textbf{BADGE}, designed for this purpose using LLM. Our method consists of two main phases: Report Generation and Report Evaluation. Initially, badminton-related data is processed by the LLM, which then generates a detailed report of the match. We tested different Input Data Types, In-Context Learning (ICL), and LLM, finding that GPT-4 performs best when using CSV data type and the Chain of Thought prompting. Following report generation, the LLM evaluates and scores the reports to assess their quality. Our comparisons between the scores evaluated by GPT-4 and human judges show a tendency to prefer GPT-4 generated reports. Since the application of LLM in badminton reporting remains largely unexplored, our research serves as a foundational step for future advancements in this area. Moreover, our method can be extended to other sports games, thereby enhancing sports promotion. For more details, please refer to \href{https://github.com/AndyChiangSH/BADGE}{https://github.com/AndyChiangSH/BADGE}.

\end{abstract}

\section{Introduction}

\begin{figure}
  \centering
  \includegraphics[width=\linewidth]{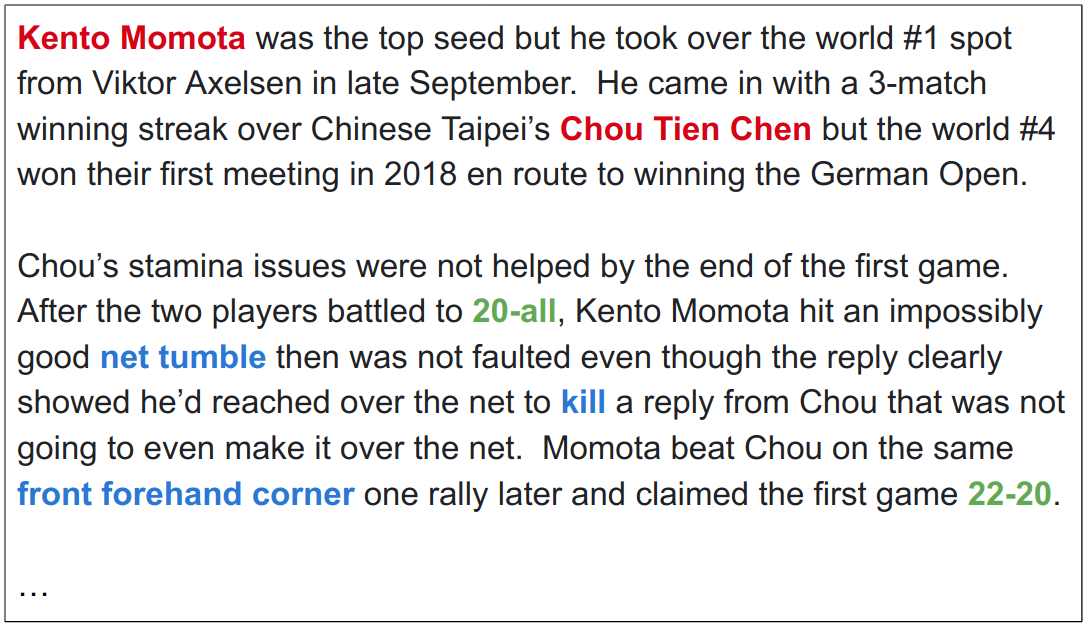}
  \caption{The example of the badminton report, where red is the player name, green is the game score, and blue is the ball type.}
  \label{fig:badminton_report}
\end{figure}

Badminton, as one of the most popular racket sports globally, demands a nuanced understanding of gameplay dynamics, player strategies, and match outcomes. However, manual analysis can be subjective and time-consuming. Therefore, we aim to automate the process of report generation, thereby facilitating faster insights extraction and broader accessibility to game analysis.
In recent years, the advent of Large Language Models (LLM) has revolutionized Natural Language Processing (NLP) tasks across various domains, ranging from text generation to language understanding \cite{ChatGPT}. Among these cutting-edge models, GPT-3.5 stands out as a widely used and publicly available tool, capable of generating coherent and contextually relevant text based on input prompts. In this paper, we explore its application in the domain of badminton game analysis, particularly focusing on the generation of comprehensive game reports, as shown in Figure \ref{fig:badminton_report}, using datasets derived from badminton matches.

In this paper, we seek to address several key research questions: How does the performance of GPT-3.5 compare across different In-Context Learning methods in the context of badminton game report generation? What are the strengths and limitations of using structured (CSV files) versus unstructured (Question-Answer pairs) input data for prompting the model? To what extent can generated reports capture the nuances of badminton gameplay, player strategies, and match outcomes compared to manually crafted reports?

The primary objective of this study is twofold. Firstly, to investigate the performance of different In-Context Learning methods and Input Data Types in enhancing the quality of generated badminton game reports. Secondly, to quantify badminton reports and compare different generation methods in order to identify the optimal approach.


By answering these questions, we aim to contribute valuable insights into the feasibility and effectiveness of employing LLMs, for automated game analysis in the realm of badminton, and provide insights into the shift of human preferences on how reports are created, paving the way for enhanced report generation and evaluation.

\section{Related Works}

\subsection{Badminton Dataset}
The current state of sports report generation using Large Language Models (LLMs) leverages the power of artificial intelligence to produce detailed, accurate, and engaging content. These models are capable of analyzing vast amounts of real-time data, including scores, player statistics, and game highlights, to generate comprehensive reports and summaries. They can craft narratives that capture the excitement and nuances of sporting events, providing insights and commentary akin to human sports journalists. The integration of LLMs in sports journalism represents a significant leap forward, enhancing both the efficiency and richness of sports coverage.

Taking in the above, we consider the following requirements for our base dataset used to generate relevant input prompts: (1) relating to the field of badminton, and (2) providing a wide spread of information outside of the game itself, such as tournament title, player names, location and so on, that are useful to generate comprehensive reports. Thus we turn to ShuttleSet \cite{wang2023shuttleset}, introduced as a meticulously curated stroke-level singles dataset designed for facilitating in-depth tactical analysis in badminton. This dataset, comprising human-annotated match data, provides a granular perspective on player performance and strategic decision-making during singles matches. By capturing stroke-level details such as shot types, placement, and rally dynamics, ShuttleSet enables researchers to delve into the intricacies of badminton gameplay and extract actionable insights for players, coaches, and analysts.

The ShuttleSet dataset encompasses a diverse range of singles matches, featuring players of varying skill levels and playing styles. Each match in the dataset is meticulously annotated to capture crucial aspects of gameplay, including shot trajectories, rally duration, and point outcomes. Moreover, the dataset includes contextual information such as player names, match settings, and tournament context, enriching the analytical capabilities and applicability of the dataset in diverse research settings.

Utilizing ShuttleSet, researchers have the opportunity to explore a multitude of research questions related to badminton tactical analysis, player performance evaluation, and strategic decision-making. By leveraging the detailed stroke-level annotations provided in the dataset, researchers can gain valuable insights into player strategies, tactical patterns, and match dynamics, ultimately enhancing our understanding of the sport and informing coaching methodologies and training regimens.


\subsection{Generation with LLM}


Our approach draws inspiration from In-Context Learning frameworks \cite{dong2022survey}, emphasizing the role of contextual information and tailored prompts. Recognizing the importance of roles for In-Context Learning demonstrations \cite{min2022rethinking}, for their potential impact on enhancing narrative coherence and content relevance. We acknowledge the advancements in prompting engineering, such as Zero-shot, One-shot, Few-shot \cite{brown2020language}, Chain of Thought \cite{wei2022chain} and automatic prompt generation mechanisms \cite{zhang2022automatic} in facilitating efficient and effective narrative construction. Leveraging the exploration of self-consistency mechanisms \cite{wang2023selfconsistency}, our method aims to elicit coherent narratives that capture the essence of badminton gameplay. We also consider the significance of deliberate problem-solving strategies, as proposed in the "Tree of Thoughts" framework \cite{NEURIPS2023_271db992}, to guide the generation process toward producing insightful reports.

Informed by a comprehensive review of the recent work mentioned above, we synthesized insights from various methodologies of prompting, including Zero-shot, One-shot, Few-shot, Chain of Thought, Auto Chain of Thought, and Tree of Thought to come up with suitable prompts for report generation, seeking to enhance the coherence and depth of generated badminton game reports, aligning with the nuances of match dynamics and player performances.

\subsection{Evaluation with LLM}



To evaluate the generated reports, we surveyed several evaluation methods. Sai et al.'s survey \cite{sai2020survey} provides an overview of various evaluation metrics for Natural Language Generation (NLG) systems, offering a broad perspective on their applicability, or lack thereof, within the rapidly evolving field of NLG. Fu et al.'s work \cite{fu2023gptscore} introduces GPTScore, a flexible method for evaluating NLG systems, tested on a multitude of different LLM structures and sizes, to emphasize its adaptability of diverse evaluation criteria and domains. Wang et al.'s study \cite{wang2023chatgpt} presents a preliminary examination of ChatGPT's effectiveness as an NLG evaluator, highlighting its strengths and weaknesses through empirical analysis of five NLG meta-evaluation datasets (including summarization, story generation and data-to-text tasks). Liu et al. proposed the G-Eval framework \cite{liu2023geval}, which encompasses chain-of-thought and weighting techniques for assessing the coherence, consistency, and fluency of news summaries.

After considering these methods, we find G-Eval sufficient and applicable, ultimately deciding to utilize their framework, since empirical evidence show results of its evaluation better aligning with human judgments. By systematically evaluating the generated reports against human-authored references and benchmarking against established evaluation criteria, we aim to gain insights into the performance characteristics of our proposed generation method and identify areas for improvement.


\section{Methods}


\subsection{Overview}

\begin{figure}
  \centering
  \includegraphics[width=\linewidth]{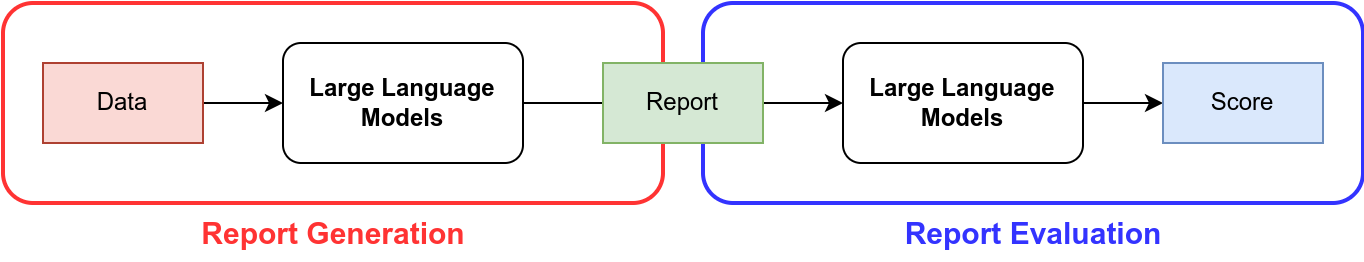}
  \caption{The overview of our proposed framework, \textbf{BADGE}}
  \label{fig:overview}
\end{figure}

Figure \ref{fig:overview} presents an overview of our proposed framework, \textbf{BADGE}. This framework separates the whole process into two distinct stages: (1) Report Generation and (2) Report Evaluation. During the first stage, the input consists of badminton data retrieved from ShuttleSet \cite{wang2023shuttleset}. This data is then processed by the LLM to generate a badminton report. In the second stage, the LLM evaluates the report generated in the previous stage,  resulting in a corresponding evaluation score.

\subsection{Report Generation}

\begin{figure}
  \centering
  \includegraphics[width=\linewidth]{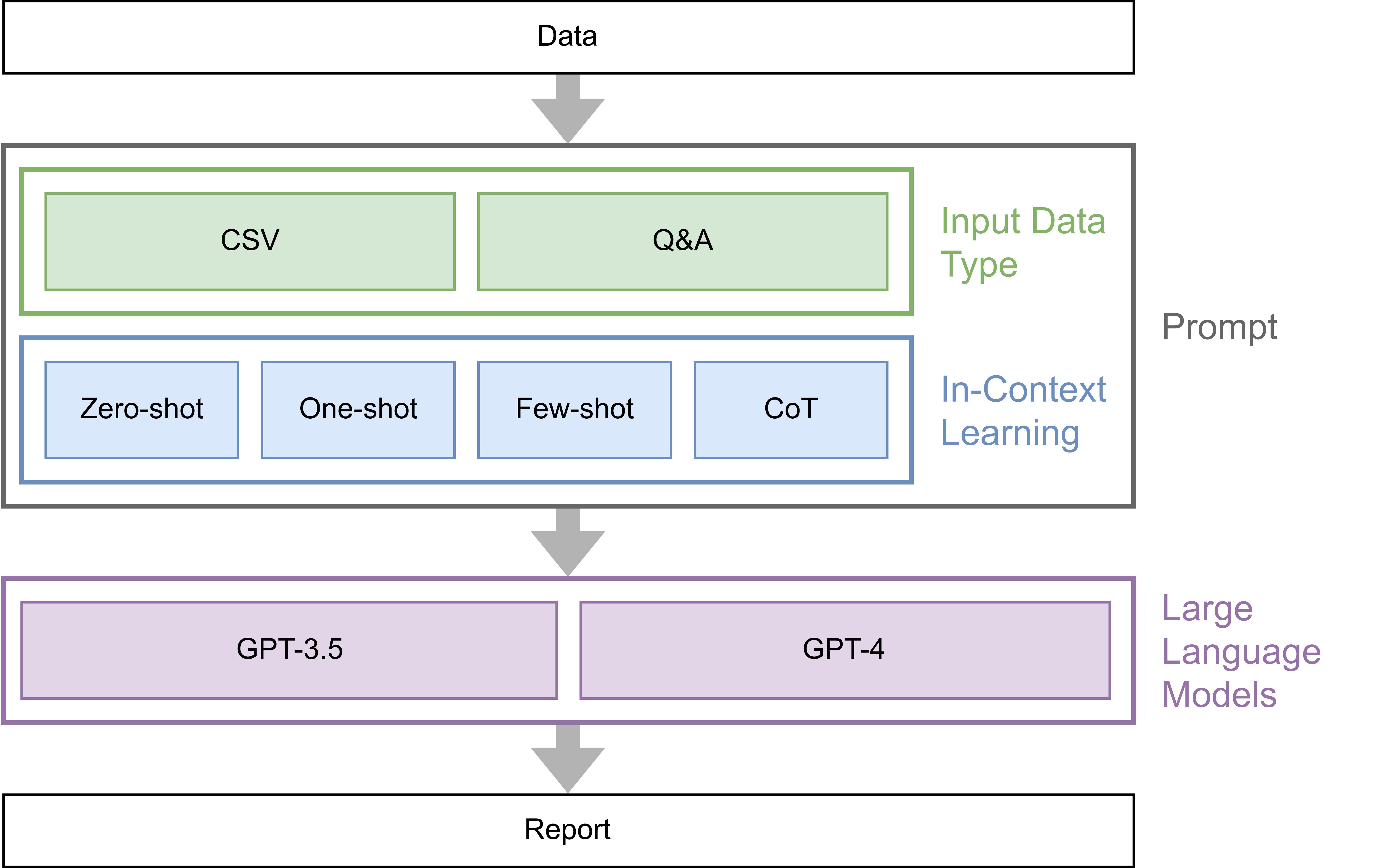}
  \caption{The flowchart of Report Generation}
  \label{fig:generation}
\end{figure}


For report generation, we employ diverse Input Data Types, methods of In-Context Learning (ICL), and Large Language Models (LLM). The flowchart of the Report Generation is shown in Figure \ref{fig:generation}.

\subsubsection{Input Data Type}


To compare the differences between structured and unstructured data, we utilize two distinct input data types to represent the badminton game: CSV and Q\&A. CSV, an acronym for "Comma-Separated Values," denotes a straightforward and widely adopted file format for storing tabular data, such as spreadsheets or databases. In a CSV file, each line represents a row of data, with the features within each row separated by commas. This format represents the rally-level data of the badminton game. On the other hand, Q\&A, which stands for "Question and Answer," involves designing eight questions pertinent to a badminton set. A rule-based Python code is responsible for computing the answer to each question and then filling the answers into the predefined template. This format represents the set-level data of the badminton game. Examples illustrating CSV and Q\&A formats are provided below:\\


\begin{quote}
{\fontfamily{cmss}\selectfont
\textbf{CSV:}\\
win\_point\_player, win\_reason, ball\_types, lose\_reason, roundscore\_A, roundscore\_B\\
Ratchanok Intanon, opponent goes out of bounds, lob, goes out of bounds, 0, 1\\
An Se Young, opponent hits the net, push, hits the net, 1, 1\\
Ratchanok Intanon, wins by landing, smash, opponent wins by landing, 1, 2\\
...\\
}
\end{quote}

\begin{quote}
{\fontfamily{cmss}\selectfont
\textbf{Q\&A:}\\
Q1: Which player won the game? How many points did the winner get?\\
A1: An Se Young won the game with 22 points.\\
Q2: Which player lost the game? How many points did the loser get?\\
A2: Ratchanok Intanon lost the game with 20 points.\\
...\\
}
\end{quote}

\subsubsection{In-Context Learning (ICL)}


To facilitate In-Context Learning, we design four distinct prompt types, drawing inspiration from existing literature \cite{dong2022survey}: Zero-shot, One-shot, Few-shot \cite{brown2020language}, and Chain of Thought (CoT) \cite{wei2022chain}.  Zero-shot prompts involve no illustrative examples during inference. One-shot prompts provide a single example, while Few-shot prompts offer a limited number of examples at inference time. Chain of Thought (CoT) is a technique that empowers LLM to tackle complex reasoning tasks by thinking them step by step. It essentially breaks down the problem into smaller, more manageable chunks for the LLM to process. The prompts of In-Context Learning are shown below:\\

\begin{quote}
{\fontfamily{cmss}\selectfont
\textbf{Zero-shot:}\\
You are a reporter for badminton games.\\
...\\
}
\end{quote}

\begin{quote}
{\fontfamily{cmss}\selectfont
\textbf{One-shot:}\\
You are a reporter for badminton games.\\
...\\
I give you an example report as a reference:\\
Example:\\
...\\
}
\end{quote}

\begin{quote}
{\fontfamily{cmss}\selectfont
\textbf{Few-shot:}\\
You are a reporter for badminton games.\\
...\\
I give you some example reports as reference:\\
Example 1:\\
...\\
Example 2:\\
...\\
}
\end{quote}

\begin{quote}
{\fontfamily{cmss}\selectfont
\textbf{CoT:}\\
You are a reporter for badminton games.\\
...\\
Let's think step by step:\\
1. Read the CSV table carefully and understand this badminton game.\\
2. ...\\
}
\end{quote}

\subsubsection{Large Language Models (LLM)}


To compare the different LLMs for report generation, we utilize GPT-3.5 (GPT-3.5-turbo-0125) \cite{ChatGPT} and GPT-4 (GPT-4-turbo-2024-04-09) \cite{achiam2023gpt} to generate the badminton reports. Both GPT-3.5 and GPT-4 are accessed through the \href{https://openai.com/blog/openai-api}{OpenAI API}.

\subsection{Report Evaluation}


Evaluating the quality of texts generated by Natural Language Generation (NLG) systems presents challenges in automated measurement. Furthermore, conventional reference-based metrics like BLEU \cite{papineni2002bleu} and ROUGE \cite{lin2004rouge} have demonstrated limited correlation with human judgments, particularly in tasks demanding creativity and diversity. Consequently, recent research advocates for leveraging LLMs as reference-free metrics for NLG evaluation \cite{wang2023chatgpt} \cite{liu2023geval}. In our study, we introduce two evaluation methodologies: GPT-4 Evaluation and Human Evaluation.

\subsubsection{GPT-4 Evaluation}

\begin{figure}
  \centering
  \includegraphics[width=\linewidth]{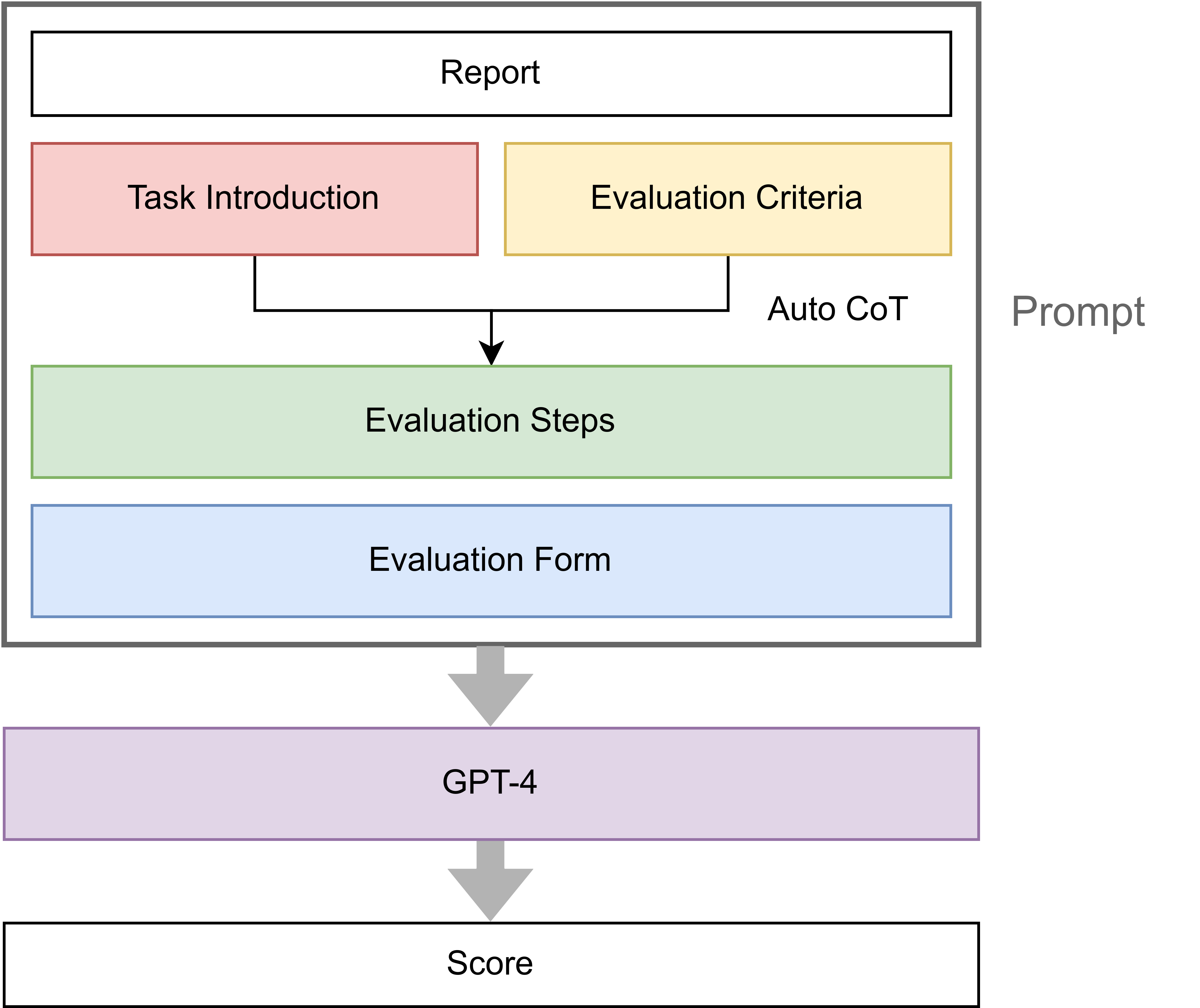}
  \caption{The flowchart of GPT-4 Evaluation}
  \label{fig:evaluation}
\end{figure}


We follow the framework presented in the G-EVAL paper \cite{liu2023geval}, with the corresponding flowchart depicted in Figure \ref{fig:evaluation}. Initially, we design the prompt for the task introduction and establish the evaluation criteria. An example of the task introduction is as follows:\\

\begin{quote}
{\fontfamily{cmss}\selectfont
\textbf{Task Introduction:}\\
You are a reviewer of the badminton reports.\\
I will give a badminton report, please follow the Evaluation Steps to score this badminton report based on the Evaluation Criteria.\\
...\\
}
\end{quote}

Our evaluation framework encompasses four criteria: coherence, consistency, excitement, and fluency. Here are the definitions for each of these evaluation criteria:\\

\begin{itemize}
{\fontfamily{cmss}\selectfont
    \item \textbf{Coherence (1-10):} means being logical and clear in thought or communication, where ideas fit together smoothly to form a unified whole.
    \item \textbf{Consistency (1-10):} refers to the quality of being steadfast, reliable, and uniform in behavior, performance, or appearance over time.
    \item \textbf{Excitement (1-10):} is a feeling of enthusiasm or thrill, often before or during an event or activity.
    \item \textbf{Fluency (1-10):} the quality of the summary in terms of grammar, spelling, punctuation, word choice, and sentence structure.\\
}
\end{itemize}


Subsequently, we will utilize the task introduction and evaluation criteria to automatically generate the evaluation steps by GPT-4. Examples of these evaluation steps are provided below:\\

\begin{quote}
{\fontfamily{cmss}\selectfont
\textbf{Evaluation Steps:}\\
1. Read for Structure and Organization: ...\\
2. Sentence-Level Analysis: ...\\
3. Overall Coherence Assessment: ...\\
}
\end{quote}


Finally, we integrate the task introduction, evaluation criteria, evaluation steps, badminton report, and evaluation form into the input prompt. GPT-4 will then assign a score on a scale of 1 to 10, where 1 represents the lowest and 10 denotes the highest, based on the specified evaluation criteria. Each evaluation criterion is assessed individually during the evaluation process.

\subsubsection{Human Evaluation}


To compare the correlation between evaluations by GPT-4 and humans, we conduct human evaluations on our badminton reports. For the human evaluation, we prepared a form containing three badminton reports authored by GPT-3.5, GPT-4, and humans, respectively. Subsequently, evaluators will assign scores to each badminton report based on four evaluation criteria: coherence, consistency, excitement, and fluency. Additionally, evaluators will attempt to identify the author of each report. Finally, we will calculate the average scores assigned by the evaluators and compare them with the scores evaluated by GPT-4.

\section{Experiments}


\subsection{Dataset}


We sample 10 badminton games spanning the years 2018 to 2021 from ShuttleSet \cite{wang2023shuttleset}. Among these games, 5 pertain to men's singles, while the remaining 5 feature women's singles matches. Each game comprises 2 or 3 sets, with each set containing 30 columns of features. However, for the sake of simplification, we only extract the 6 most crucial columns, which include win\_point\_player, win\_reason, lose\_reason, ball\_types, roundscore\_A, and roundscore\_B.

\subsection{Result for Input Data Type}

\begin{table}[]
\resizebox{\columnwidth}{!}{%
\begin{tabular}{cccccc}
\toprule
\textbf{Data Type + ICL} & \textbf{Coherence} & \textbf{Consistency} & \textbf{Excitement} & \textbf{Fluency} & \textbf{Avg.}  \\
\midrule
\textit{CSV + zero-shot}   & 8.2                & \textit{7.5}         & 7.9                 & 8.8              & \textit{8.100} \\
CSV + one-shot             & 8.4                & 8.3                  & 7.8                 & 8.8              & 8.325          \\
CSV + few-shot             & 8.3                & 9.0                  & 7.7                 & 8.7              & 8.425          \\
\textbf{CSV + CoT}         & 8.4                & \textbf{9.2}         & \textbf{8.0}        & \textbf{8.9}     & \textbf{8.625} \\
\midrule
Q\&A + zero-shot             & \textit{7.9}       & 8.6                  & \textit{7.3}        & 8.7              & 8.125          \\
Q\&A + one-shot              & \textbf{8.6}       & 8.4                  & 7.4                 & 8.8              & 8.300          \\
Q\&A + few-shot              & 8.3                & 8.5                  & 7.5                 & 8.6              & 8.225          \\
Q\&A + CoT                   & \textit{7.9}       & 8.7                  & 7.4                 & \textit{8.5}     & 8.125         \\
\bottomrule
\end{tabular}%
}
\caption{The result of GPT-4 evaluation for reports with different input data types and ICL, where \textbf{bold} denotes the best result and \textit{italics} indicates the worst result.}
\label{tab:ICL}
\end{table}

\begin{figure}
  \centering
  \includegraphics[width=\linewidth]{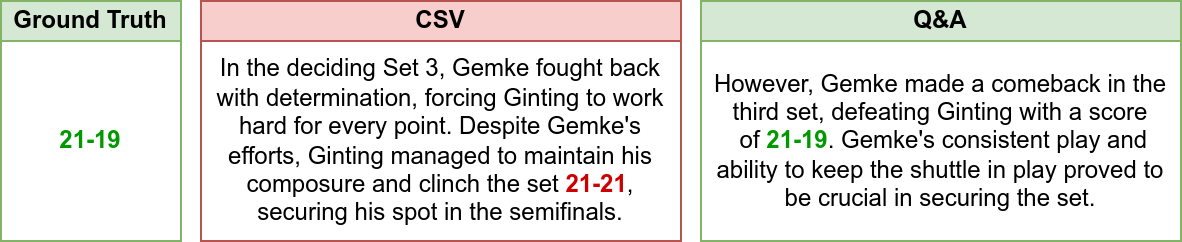}
  \caption{The example of generated reports with CSV and Q\&A data types. Green indicates the correct score, while red indicates an incorrect score.}
  \label{fig:data_type}
\end{figure}


To compare the reports generated with different data types and ICL, we generate reports using two data types and four ICL techniques with GPT-3.5. Subsequently, all reports are evaluated by GPT-4, with the scores representing the average score for each evaluation criterion across 10 games. The results are presented in Table \ref{tab:ICL}.


As observed, reports utilizing the CSV data type exhibit slightly better performance in terms of consistency, excitement, and fluency compared to those employing the Q\&A data type. However, it is notable that reports with the CSV data type are more prone to hallucinations. For example, referring to Figure \ref{fig:data_type}, while the ground truth score is 21-19, the score in the report with the Q\&A data type is correct. Conversely, the score in the report with the CSV data type is 21-21, which is incorrect.

\subsection{Result for In-Context Learning (ICL)}


In Table \ref{tab:ICL}, we observed that Chain of Thought demonstrated the best overall performance with an approximately 0.2 improvement over the few-shot on the CSV data type, followed by one-shot, and zero-shot, in descending order. Therefore, we speculate that Chain of Thought divides the task into multiple smaller tasks, enabling the LLM to generate better reports step by step. We also discovered that increasing the number of demonstrations improves the evaluation scores, proving the effectiveness of demonstrations. However, a similar pattern was not evident for the Q\&A data type. Consequently, we hypothesize that the data type may also be a factor influencing the performance of ICL.

\subsection{Result for Large Language Models (LLM)}


To compare the quality of reports generated by GPT-3.5, GPT-4, and human writers, we generate reports using GPT-3.5 and GPT-4, and collect human-written reports from the Internet. All reports are then evaluated by GPT-4, and the scores represent the average score for each evaluation criterion across 10 games. The experimental results are presented in Table \ref{tab:GPT-4_evaluation}.


We observe that reports generated by GPT-4 exhibit the highest performance, whereas human-written reports receive the lowest scores across all four evaluation criteria.

\begin{table}[]
\resizebox{\columnwidth}{!}{%
\begin{tabular}{@{}cccccc@{}}
\toprule
\textbf{Writer} & \textbf{Coherence} & \textbf{Consistency} & \textbf{Excitement} & \textbf{Fluency} & \textbf{Avg.}  \\ \midrule
\textit{Human}  & \textit{7.5}       & \textit{8.9}         & \textit{6.8}        & \textit{8.5}     & \textit{7.925} \\
GPT-3.5         & 8.4                & 9.2                  & 8.0                 & 8.9              & 8.625          \\
\textbf{GPT-4}  & \textbf{8.6}       & \textbf{9.4}         & \textbf{8.2}        & \textbf{9.1}     & \textbf{8.825} \\ \bottomrule
\end{tabular}%
}
\caption{The result of GPT-4 evaluation for reports written by humans, GPT-3.5, and GPT-4, where \textbf{bold} denotes the best result and \textit{italics} indicates the worst result.}
\label{tab:GPT-4_evaluation}
\end{table}

\begin{table}[]
\resizebox{\columnwidth}{!}{%
\begin{tabular}{@{}cccccc@{}}
\toprule
\textbf{Writer}  & \textbf{Coherence} & \textbf{Consistency} & \textbf{Excitement} & \textbf{Fluency} & \textbf{Avg.}  \\ \midrule
Human            & 7.6                & 7.5                  & 6.9                 & 7.8              & 7.450           \\
\textit{GPT-3.5} & \textit{6.5}       & \textit{7.3}         & \textit{5.2}        & \textit{6.4}     & \textit{6.350}  \\
\textbf{GPT-4}   & \textbf{8.3}       & \textbf{8.2}         & \textbf{8.0}          & \textbf{8.4}     & \textbf{8.225} \\ \bottomrule
\end{tabular}%
}
\caption{The result of human evaluation for reports written by humans, GPT-3.5, and GPT-4, where \textbf{bold} denotes the best result and \textit{italics} indicates the worst result.}
\label{tab:human_evaluation}
\end{table}

\subsection{Result for Human Evaluation}

\begin{figure}
  \centering
  \includegraphics[width=\linewidth]{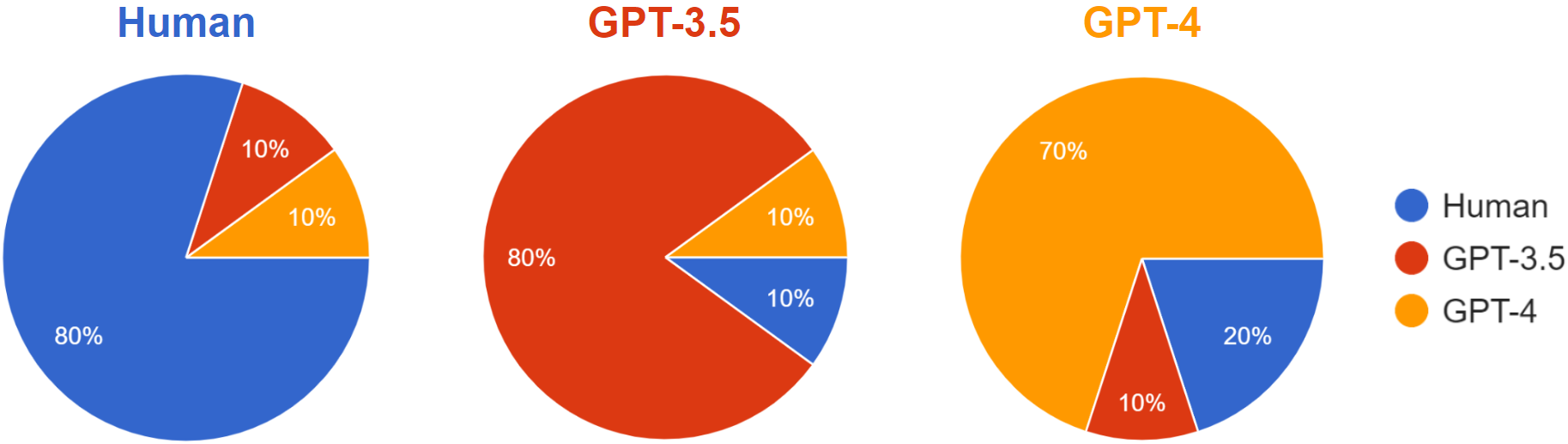}
  \caption{The accuracy of guessing who wrote the report.}
  \label{fig:guess}
\end{figure}


The experimental results are presented in Table \ref{tab:human_evaluation}. Most evaluators rated the report generated by GPT-4 as the best, while preferring the human-written report over the one generated by GPT-3.5. This finding contradicts the evaluation by GPT-4, where GPT-3.5 outperformed humans. This bias aligns with observations from the G-EVAL paper \cite{liu2023geval}, which compared the GPT-4 and human evaluation and found that G-EVAL prefers the output generated by LLMs. Additionally, the Pearson product-moment correlation coefficient between the GPT-4 evaluation and human evaluation is calculated to be 0.333, indicating a small positive correlation between the two evaluations.


Figure \ref{fig:guess} is the pie chart illustrating the percentage of correct guesses for each report. The accuracy rates are as follows: human reports 80\%, GPT-3 80\%, and GPT-4 70\%. These results indicate that evaluators can readily discern the author of the report in most cases, suggesting differences in the stylistic characteristics between reports authored by humans and those generated by LLMs. The examples of the reports can be found in the Appendix \ref{sec:example}.

\section{Limitations \& Future Works}






There are some limitations and future work in our framework.
Firstly, badminton report generation is a relatively unexplored topic in the research field, leaving us without other baselines for comparison. Our future work could involve constructing a benchmark (comprising dataset and evaluation metrics) to inspire and facilitate further research.

Secondly, we currently lack a quantitative method to measure the occurrence of hallucinations in the reports. In the future, employing a Q\&A model to extract answers from reports and comparing them with the answers obtained from a rule-based Python code could offer a means to calculate the accuracy rate.

Finally, the bias that GPT-4 prefers the reports generated by LLM may lead to unfair evaluation. Exploring solutions to this issue represents a promising direction for future research.

\section{Conclusion}

In conclusion, our work marks a pioneering venture into badminton report generation and evaluation. Our innovative framework, \textbf{BADGE}, separates the process into two stages: Report Generation and Report Evaluation. Initially, badminton data sourced from ShuttleSet serves as input, processed by the LLM to generate the reports to describe the badminton game. Subsequently, in the evaluation stage, the LLM assesses the reports, yielding corresponding scores. Our experiments encompass comparisons across different Input Data Types, In-Context Learning (ICL), and Large Language Models (LLMs). We found that reports generated by GPT-4 with CSV and Chain of Thought exhibit the best performance. Moreover, we compared the scores evaluated by GPT-4 and humans, revealing a bias where GPT-4 favors reports generated by LLMs. Despite existing limitations, our work sets the stage for future advancements in badminton report generation and evaluation, potentially paving the way for research and innovation in this field.

\section*{Acknowledgments}

This work was supported by the Ministry of Science and Technology of Taiwan under Grants 113-2425-H-A49-001.

\bibliographystyle{named}
\bibliography{ijcai24}

\begin{thebibliography}{}

\bibitem[\protect\citeauthoryear{Achiam \bgroup \em et al.\egroup }{2023}]{achiam2023gpt}
Josh Achiam, Steven Adler, Sandhini Agarwal, Lama Ahmad, Ilge Akkaya, Florencia~Leoni Aleman, Diogo Almeida, Janko Altenschmidt, Sam Altman, Shyamal Anadkat, et~al.
\newblock Gpt-4 technical report.
\newblock {\em arXiv preprint arXiv:2303.08774}, 2023.

\bibitem[\protect\citeauthoryear{Brown \bgroup \em et al.\egroup }{2020}]{brown2020language}
Tom Brown, Benjamin Mann, Nick Ryder, Melanie Subbiah, Jared~D Kaplan, Prafulla Dhariwal, Arvind Neelakantan, Pranav Shyam, Girish Sastry, Amanda Askell, et~al.
\newblock Language models are few-shot learners.
\newblock {\em Advances in neural information processing systems}, 33:1877--1901, 2020.

\bibitem[\protect\citeauthoryear{Dong \bgroup \em et al.\egroup }{2022}]{dong2022survey}
Qingxiu Dong, Lei Li, Damai Dai, Ce~Zheng, Zhiyong Wu, Baobao Chang, Xu~Sun, Jingjing Xu, and Zhifang Sui.
\newblock A survey on in-context learning.
\newblock {\em arXiv preprint arXiv:2301.00234}, 2022.

\bibitem[\protect\citeauthoryear{Fu \bgroup \em et al.\egroup }{2023}]{fu2023gptscore}
Jinlan Fu, See-Kiong Ng, Zhengbao Jiang, and Pengfei Liu.
\newblock Gptscore: Evaluate as you desire, 2023.

\bibitem[\protect\citeauthoryear{Lin}{2004}]{lin2004rouge}
Chin-Yew Lin.
\newblock Rouge: A package for automatic evaluation of summaries.
\newblock In {\em Text summarization branches out}, pages 74--81, 2004.

\bibitem[\protect\citeauthoryear{Liu \bgroup \em et al.\egroup }{2023}]{liu2023geval}
Yang Liu, Dan Iter, Yichong Xu, Shuohang Wang, Ruochen Xu, and Chenguang Zhu.
\newblock G-eval: Nlg evaluation using gpt-4 with better human alignment, 2023.

\bibitem[\protect\citeauthoryear{Min \bgroup \em et al.\egroup }{2022}]{min2022rethinking}
Sewon Min, Xinxi Lyu, Ari Holtzman, Mikel Artetxe, Mike Lewis, Hannaneh Hajishirzi, and Luke Zettlemoyer.
\newblock Rethinking the role of demonstrations: What makes in-context learning work?, 2022.

\bibitem[\protect\citeauthoryear{OpenAI}{2022}]{ChatGPT}
OpenAI.
\newblock Introducing chatgpt.
\newblock \url{https://openai.com/blog/chatgpt}, 2022.

\bibitem[\protect\citeauthoryear{Papineni \bgroup \em et al.\egroup }{2002}]{papineni2002bleu}
Kishore Papineni, Salim Roukos, Todd Ward, and Wei-Jing Zhu.
\newblock Bleu: a method for automatic evaluation of machine translation.
\newblock In {\em Proceedings of the 40th annual meeting of the Association for Computational Linguistics}, pages 311--318, 2002.

\bibitem[\protect\citeauthoryear{Sai \bgroup \em et al.\egroup }{2020}]{sai2020survey}
Ananya~B. Sai, Akash~Kumar Mohankumar, and Mitesh~M. Khapra.
\newblock A survey of evaluation metrics used for nlg systems, 2020.

\bibitem[\protect\citeauthoryear{Wang \bgroup \em et al.\egroup }{2023a}]{wang2023chatgpt}
Jiaan Wang, Yunlong Liang, Fandong Meng, Zengkui Sun, Haoxiang Shi, Zhixu Li, Jinan Xu, Jianfeng Qu, and Jie Zhou.
\newblock Is chatgpt a good nlg evaluator? a preliminary study, 2023.

\bibitem[\protect\citeauthoryear{Wang \bgroup \em et al.\egroup }{2023b}]{wang2023shuttleset}
Wei-Yao Wang, Yung-Chang Huang, Tsi-Ui Ik, and Wen-Chih Peng.
\newblock Shuttleset: A human-annotated stroke-level singles dataset for badminton tactical analysis.
\newblock In {\em Proceedings of the 29th ACM SIGKDD Conference on Knowledge Discovery and Data Mining}, pages 5126--5136, 2023.

\bibitem[\protect\citeauthoryear{Wang \bgroup \em et al.\egroup }{2023c}]{wang2023selfconsistency}
Xuezhi Wang, Jason Wei, Dale Schuurmans, Quoc Le, Ed~Chi, Sharan Narang, Aakanksha Chowdhery, and Denny Zhou.
\newblock Self-consistency improves chain of thought reasoning in language models, 2023.

\bibitem[\protect\citeauthoryear{Wei \bgroup \em et al.\egroup }{2022}]{wei2022chain}
Jason Wei, Xuezhi Wang, Dale Schuurmans, Maarten Bosma, Fei Xia, Ed~Chi, Quoc~V Le, Denny Zhou, et~al.
\newblock Chain-of-thought prompting elicits reasoning in large language models.
\newblock {\em Advances in neural information processing systems}, 35:24824--24837, 2022.

\bibitem[\protect\citeauthoryear{Yao \bgroup \em et al.\egroup }{2023}]{NEURIPS2023_271db992}
Shunyu Yao, Dian Yu, Jeffrey Zhao, Izhak Shafran, Tom Griffiths, Yuan Cao, and Karthik Narasimhan.
\newblock Tree of thoughts: Deliberate problem solving with large language models.
\newblock In A.~Oh, T.~Neumann, A.~Globerson, K.~Saenko, M.~Hardt, and S.~Levine, editors, {\em Advances in Neural Information Processing Systems}, volume~36, pages 11809--11822. Curran Associates, Inc., 2023.

\bibitem[\protect\citeauthoryear{Zhang \bgroup \em et al.\egroup }{2022}]{zhang2022automatic}
Zhuosheng Zhang, Aston Zhang, Mu~Li, and Alex Smola.
\newblock Automatic chain of thought prompting in large language models, 2022.

\end{thebibliography}


\appendix

\section{Example} \label{sec:example}

\subsection{Report Written by Human}

{\fontfamily{cmss}\selectfont
Kento Momota was the top seed but he took over the world \#1 spot from Viktor Axelsen in late September.  He came in with a 3-match winning streak over Chinese Taipei’s Chou Tien Chen but the world \#4 won their first meeting in 2018 en route to winning the German Open.

Chou Tien Chen has been one of the busiest top ten players in the past few months, if not the past year.  He has played 7 tournaments (plus the Asian Games team event) in the last 12 weeks and in the last 8 weeks, he has reached 5 semi-finals, converting 3 of those into appearances in the finals.

Chou’s stamina issues were not helped by the end of the first game.  After the two players battled to 20-all, Kento Momota (pictured right) hit an impossibly good net tumble then was not faulted even though the reply clearly showed he’d reached over the net to kill a reply from Chou that was not going to even make it over the net.  Momota beat Chou on the same front forehand corner one rally later and claimed the first game 22-20.

Chou bounced back to win the second game – despite being unable to convert on his first 4 game point opportunities – but then Momota surged ahead late in the decider to cap off his victory.  Momota thus claimed his 5th title of the year.  As two of these were the World and the Asian Championships, he has been accumulating slightly fewer points on the BWF World Tour and is now the only world \#1 who did not also finish the weekend on top of the Race to Guangzhou standings.

All of the finalists will now head to Paris for the French Open Super 750.  Chou Tien Chen and Zheng/Huang have put themselves in the position such that they are assured of qualifying for the World Tour Finals as long as they merely step on court for their first round matches in Paris, Fuzhou, and Hong Kong.
}

\subsection{Report Written by GPT-3.5}

{\fontfamily{cmss}\selectfont
In the Denmark Open 2018 Finals, Kento MOMOTA faced off against CHOU Tien Chen in an intense badminton showdown.

In the first set, Kento MOMOTA emerged victorious with a score of 22-20 against CHOU Tien Chen. MOMOTA secured 7 points with wins by landing, while CHOU Tien Chen struggled with 7 points lost due to opponent wins by landing. MOMOTA excelled with 2 points from rush shots, while CHOU Tien Chen faltered with 5 points lost from return nets. MOMOTA displayed his determination by using a return net to overtake the lead at 16-15 and ultimately ended the game with a net shot.

Moving on to the second set, MOMOTA continued his winning streak with a score of 21-15 against CHOU Tien Chen. MOMOTA capitalized on 8 points gained from opponents going out of bounds, while CHOU Tien Chen suffered with 8 points lost from going out of bounds as well. MOMOTA's strong performance included 5 points from smash shots, while CHOU Tien Chen struggled with 4 points lost from lobs. MOMOTA used a lob to overtake the lead at 5-4 and secured the game by forcing CHOU Tien Chen out of bounds with a push shot.

In the final set, It was CHOU Tien Chen who emerged victorious with a score of 21-16 against Kento MOMOTA. CHOU Tien Chen gained 7 points from opponents going out of bounds, while MOMOTA lost 7 points due to going out of bounds. CHOU Tien Chen excelled with 2 points from smash shots, while MOMOTA struggled with 3 points lost from lobs. CHOU Tien Chen sealed the game by forcing MOMOTA out of bounds with a lob shot.

Overall, it was a thrilling match between two skilled players, with MOMOTA showcasing his talent in the first two sets, and CHOU Tien Chen making a strong comeback in the final set to claim the Denmark Open 2018 title.
}

\subsection{Report Written by GPT-4}

{\fontfamily{cmss}\selectfont
At the gripping Denmark Open 2018 finals, Kento Momota and Chou Tien Chen battled fiercely across three sets, showcasing the best of international badminton. Kento Momota, demonstrating remarkable precision and strategy, clinched a decisive victory against Chou Tien Chen, captivating badminton enthusiasts with an exhilarating performance.

The contest began intensely in the first set, where both players exchanged close points. Momota emerged victorious with a narrow 22-20 win. Notably, he obtained 7 points through skilled plays that landed successfully into Chou's court, hinting at his dominant net-play and court coverage capabilities. Chou also demonstrated skill but yielded 7 points to Momota under the same condition, with his returns just falling short. The critical juncture came when Momota used a masterful return net to gain a slight edge at 16:15, showing his calm under pressure. He sealed the set strategically with a net shot that decisively ended the game.

In the second set, Momota increased his momentum, securing a 21-15 victory. Here, his ability to force errors became evident, as he earned 8 points when Chou's responses went out of bounds. Momota's offensive was spearheaded by powerful smashes, recording 5 points, displaying his aggressive yet precise playstyle. Again, he skillfully overtook a close score of 5:4 using a lob, showcasing versatile skills to mesmerize the audience and outmaneuver Chou. He finalized the set victory with a push that sent the shuttle out of Chou’s reach, marking a defining moment in the match.

In a surprising turnaround during the third set, Chou Tien Chen revitalized his gameplay to claim a 21-16 set win. Although he could not overturn the match result, he displayed resilience and skill, managing to secure points frequently when Momota's shots went beyond the bounds. Chou’s defense turned robust, particularly evident in how he ended the set, using a strategic lob that Momota pushed out, securing his only set win of the match.

Overall, the finals were a testament to both players' capabilities and the intense competitiveness of professional badminton. Kento Momota's two-set win demonstrated his championship caliber through tactical superiority and unyielding determination, leaving the audience both thrilled and eager for future contests in the world of badminton.
}

\end{document}